\pdfoutput=1

\documentclass[11pt,x11names]{article}

\usepackage[final]{ACL2023}

\usepackage{times}
\usepackage{latexsym}

\usepackage[T1]{fontenc}

\usepackage[utf8]{inputenc}
\usepackage{xcolor}
\usepackage{colortbl}
\usepackage{microtype}
\usepackage{inconsolata}
\usepackage{amsmath}

\usepackage{dsfont}
\usepackage{amsfonts}
\usepackage{amssymb}
\usepackage{pifont}
\usepackage{tabularx}
\usepackage{arydshln}

\usepackage{mathtools}
\usepackage{adjustbox}
\usepackage{multirow}
\usepackage{paralist}
\usepackage{CJKutf8}
\usepackage{booktabs}
\usepackage{nicefrac}
\usepackage{enumerate}
\usepackage{bbding} 
\usepackage{wasysym}
\usepackage{caption}
\usepackage{subcaption}
\usepackage{enumitem}
\usepackage[ruled,linesnumbered]{algorithm2e}
\SetKwInOut{Parameter}{Parameters}

\usepackage{amsthm}
\theoremstyle{definition}

\newcommand\Language[1]{\colorbox[RGB]{224,235,247}{#1}}

\newcommand\InputName[1]{\colorbox[RGB]{208,226,236}{#1}}
\newcommand\SourceLanguage[1]{\colorbox{LemonChiffon2}{#1}}
\newcommand\TargetLanguage[1]{\colorbox{Honeydew2}{#1}}
\newcommand\TaskName[1]{\colorbox{LavenderBlush2}{#1}}

\usepackage{tikz}
\usepackage[edges]{forest}
\usepackage{arydshln}
\usepackage{pgfplots}
\usepackage{makecell}
\usepackage{custom} 
\usepackage{dsfont}
\usepackage{courier}
\usepackage[most]{tcolorbox}
\definecolor{blue1}{rgb}{0.15,0.15,0.15}
\definecolor{blue2}{rgb}{0.30,0.30,0.30}
\definecolor{blue3}{rgb}{0.45,0.45,0.45}
\definecolor{blue4}{rgb}{0.60,0.60,0.60}
\definecolor{blue5}{rgb}{0.70,0.70,0.70}
\definecolor{blue6}{rgb}{0.80,0.80,0.80}
\newtcolorbox{taskbox}[2][]{
	enhanced, breakable,
	colframe=blue3!40,
	colback=blue5!5,
	arc=1mm,
	outer arc=1mm,
	fontupper=\small,
	fontlower=\small,
	coltitle=blue1,
	fonttitle=\bfseries,
	boxsep=1mm,
	left=0mm,
	right=0mm,
	top=0mm,
	bottom=0mm,
	before={\noindent},
	segmentation style={solid, blue3},
	title=#2,
	#1
}
\definecolor{mycustomcolor}{RGB}{210,226,210}
\newtcolorbox{mybox}[2][]{
    width=\columnwidth,
    colback =  gray!8,
    colframe = gray!8, boxsep=0pt,left=10pt,right=10pt,top=0pt,bottom=0pt,
    fontupper=\linespread{1.1}\selectfont,
    title=#2,#1
}

\usepackage{microtype}

\usepackage{inconsolata}

\title{\textsc{AutoCAP}: Towards Automatic Cross-lingual Alignment Planning for Zero-shot Chain-of-Thought
}

\author{
	Yongheng Zhang$^{1*}$~~
	Qiguang Chen$^{2}$\thanks{~~Equal Contribution.}~~~ 
	\textbf{Min Li}$^{1}$~~~
	\textbf{Wanxiang Che}$^{2}$~~~
    \textbf{Libo Qin}$^{1}$\thanks{\,\, Corresponding Author.}\\
	$^{1}$School of Computer Science and Engineering, Central South University, China\\
	$^{2}$Research Center for SCIR, Harbin Institute of Technology, Harbin, China  \\
	\texttt{zyhbrz@gmail.com}, \texttt{qgchen@ir.hit.edu.cn}, \texttt{lbqin@csu.edu.cn}\\
}
\begin{document}
\begin{CJK}{UTF8}{gbsn}
\maketitle
\begin{abstract}
Cross-lingual chain-of-thought can effectively complete reasoning tasks across languages, which gains increasing attention.
Recently, dominant approaches in the literature improve cross-lingual alignment capabilities by integrating reasoning knowledge from different languages. Despite achieving excellent performance, current methods still have two main challenges: (1) \textit{Manual language specification}: They still highly rely on manually selecting the languages to integrate, severely affecting their generalizability; (2) \textit{Static weight allocation}: Current methods simply integrate all languages equally. In fact, different language reasoning paths should have different weights to achieve better complementation and integration.
Motivated by this, we introduce an Automatic Cross-lingual Alignment Planning (\textsc{AutoCAP}) for zero-shot chain-of-thought to address the above challenges.
The core of \textsc{AutoCAP} consists of two components: (1) \textit{Automatic Language Selection Prompting} to guide LLMs to select appropriate languages and (2) \textit{Automatic Weight Allocation Prompting} to automatically allocate alignment weight scores to each reasoning path.
    Extensive experiments on several benchmarks reveal that \textsc{AutoCAP} achieves state-of-the-art performance, surpassing previous methods that required manual effort.

\end{abstract}

\section{Introduction}

Large language models (LLMs) have achieved substantial advancements \cite{brown2020language, chen2021evaluating, qin2024large}. Particularly noteworthy is the emergence of the Chain-of-Thought (CoT), which has further enhanced the ability of LLMs to handle complex reasoning tasks \cite{wei2022chain, wang2022self}. In addition, as globalization continues to advance, aligning representations across different languages has become an urgent issue \cite{Pires_Schlinger_Garrette_2019, Mulcaire_Kasai_Smith_2019, qin2024multilingual}. This has motivated researchers to explore Cross-lingual CoT, aiming to break down language barriers by integrating CoT from different languages \cite{qin2023cross,chai2024xcot}.

\begin{figure}[t]
	\centering
	\includegraphics[width=0.48\textwidth]{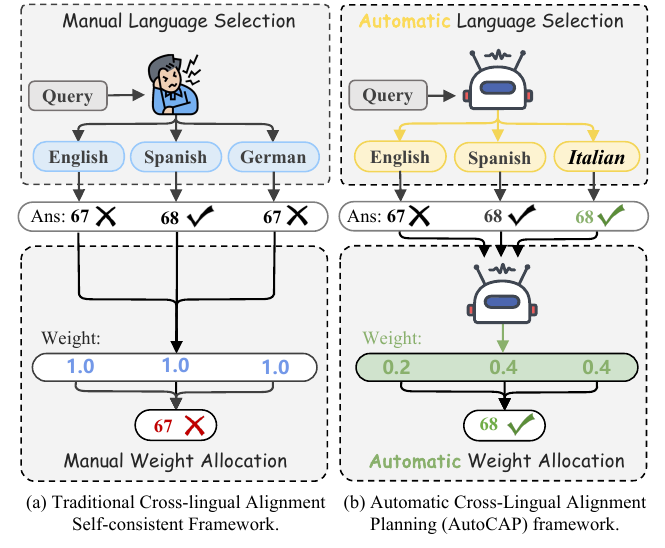}
	\caption{
	Traditional 
	Cross-lingual Self-consistent framework (a) vs. Automatic Cross-lingual Alignment Planner framework (b).
	Previous approaches require manually specifying the aligned languages (English, Spanish, and German), and assigning the same weights to these languages. In contrast, our framework (\textsc{AutoCAP}) uses the \textit{Automatic Language Selection} and \textit{Automatic Weight Allocation} to automatically select the most appropriate languages and weights.
	}
	\label{fig:intro}
\end{figure}

Specifically, \citet{shi2022language} introduce a multilingual dataset for mathematical reasoning and propose a method requiring LLMs to use English for CoT prediction and problem-solving. \citet{qin2023cross} develop a two-stage framework to aid LLMs in understanding problems across languages through manual language specification and reasoning. \citet{ranaldi2023empowering} present a cross-lingual, multi-step reasoning approach with a self-consistent prompting mechanism to enhance reasoning in multiple languages. \citet{huang2023not} introduce cross-lingual thought prompting using a generic template to enhance the reasoning capabilities and performance of LLMs on multilingual tasks. 
	In addition, recent cross-lingual CoT research has significantly improved complex reasoning across languages by aligning representations and integrating diverse linguistic reasoning paths \cite{shi2022language, tanwar2023multilingual,qin2023cross}.

Despite their improved performance, as shown in Figure~\ref{fig:intro}~(a), the current approaches still face two key challenges: 

\begin{itemize}
\item [(1)] \textit{Manual language specification}: The process continues to depend heavily on the manual selection of languages for integration, which not only expends substantial human effort but also significantly hurts its generalizability; 
\item [(2)] \textit{Static weight allocation}: 
The current methods simply integrate all languages equally, leading to sub-optimal performance.
Actually, to achieve better integration of knowledge across languages, different language reasoning paths should possess different weights relative to the query.
\end{itemize}

\begin{figure*}[t]
	\centering
	\includegraphics[width=155mm]{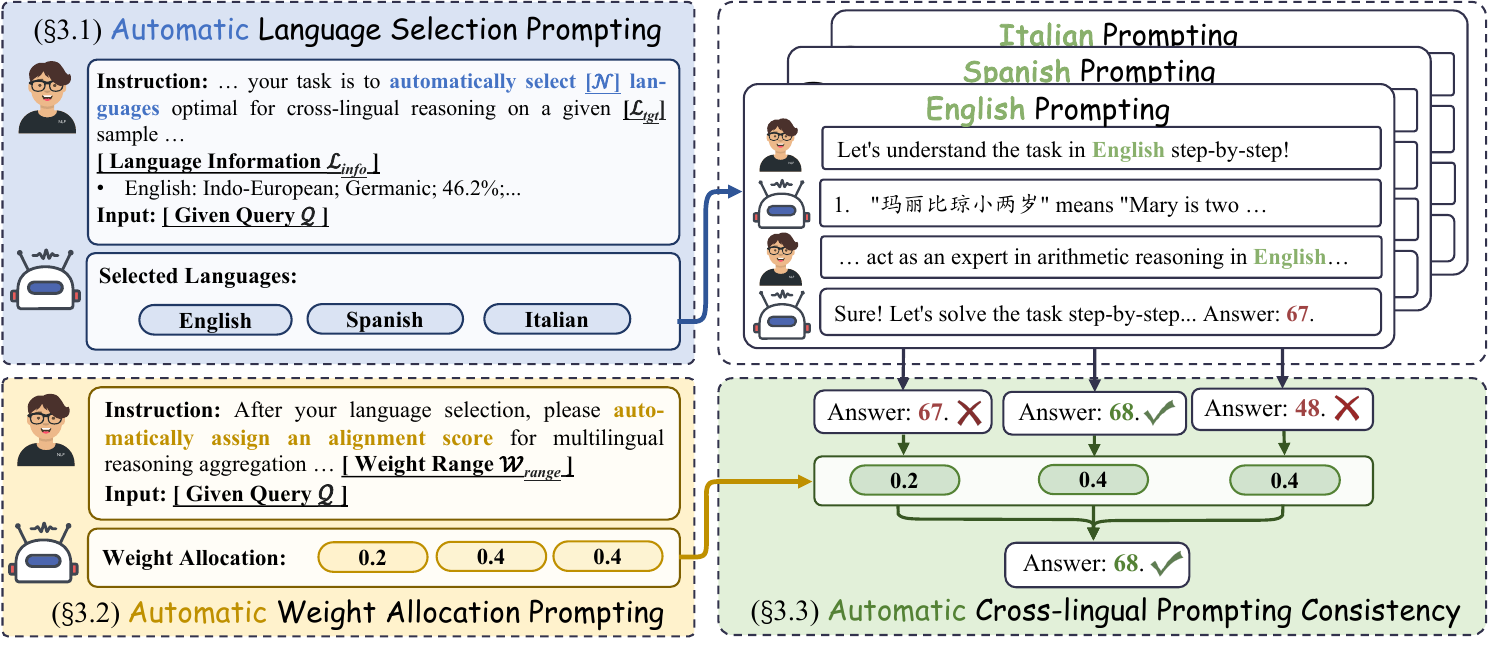}
	\caption{
		The overall workflow of \textsc{AutoCAP}, which consist of \textit{Automatic Language Selection Prompting} and \textit{Automatic Weight Allocation Prompting}.}
	\label{fig:framework}
\end{figure*}

In this paper, we introduce an automatic cross-lingual alignment planning (\textsc{AutoCAP}) framework to address the above challenges. Specifically, as shown in Figure~\ref{fig:intro}~(b), \textsc{AutoCAP} comprises two key modules: (1) \textit{Automatic Language Selection Prompting} and (2) \textit{Automatic Weight Allocation Prompting}. In more detail, \textit{Automatic Language Selection Prompting} is applied to enable LLMs to automatically select the most accurately aligned languages for reasoning for each query. After selecting the languages, \textit{Automatic Weight Allocation Prompting} is used for automatically allocating an alignment weight score to each language reasoning path. Finally, accurate reasoning answers can be obtained by integrating the CoT reasoning paths across different languages and their corresponding weight scores.

Experimental results on several benchmarks show that \textsc{AutoCAP} achieves superior performance compared to previous baselines, even surpassing previous manually selected language methods.
In addition, extensive analysis demonstrates the strong generalization ability of \textsc{AutoCAP}. 

In summary, our key contributions include:

\begin{itemize}

\item [•] We introduce Automatic Cross-lingual Alignment Planning (\textsc{AutoCAP}), which greatly alleviates the burden of manually selecting languages and weights.

\item [•] The core of \textsc{AutoCAP} comprises \textit{Automatic Language Selection Prompting} and \textit{Automatic Weight Allocation Prompting}, which achieves to automatically select the most appropriate languages and weights for cross-lingual CoT.
\item [•] Extensive experiments on several benchmarks demonstrate that \textsc{AutoCAP} surpassed the previous approaches, achieving state-of-the-art performance and exhibiting strong generalizability.

\end{itemize}

All the code will be publicly available at \url{https://github.com/BRZ911/AutoCAP}.

\section{Preliminaries}
This section outlines the preliminaries of mono-lingual chain-of-thought~($\S \ref{sec:mono-lingual}$) and cross-lingual chain-of-thought ($\S \ref{sec:cross-lingual}$).

\subsection{Mono-lingual Chain-of-Thought}\label{sec:mono-lingual}
Mono-lingual Chain-of-Thought \cite{wei2022chain} triggers LLMs to directly reason step-by-step in the source language to solve tasks. Formally, when presented with a query $\mathcal{Q}$ expressed in the source language $L_{src}$, the LLMs generate a reasoning path, which can be simplified and represented as follows:
\begin{equation}
	\mathcal{R}_{src} = {\operatorname{argmax}}\ P(\mathcal{R}|\mathcal{Q}, L_{src}),
\end{equation}
where $\mathcal{R}_{src}$ denotes the generated reasoning path with multiple steps in the source language $L_{src}$.
Following this, the LLMs produce the final results $\mathcal{Y}$, which are obtained by:
\begin{equation}
	\mathcal{Y} = {\operatorname{argmax}}\ P(y|\mathcal{Q}, L_{src}, \mathcal{R}_{src}).
\end{equation}

\subsection{Cross-lingual Chain-of-Thought}\label{sec:cross-lingual}
For better cross-lingual generalization for CoT in multilingual scenarios, \citet{qin2023cross} propose Cross-lingual Chain-of-Thought methods to align multilingual representations explicitly. Formally, given a query $\mathcal{Q}$ in source language $L_{src}$, experts \textit{{manually}} select a target language $L_{tgt}$ to serve as an anchor for cross-lingual alignment. The LLMs then generate an alignment $\mathcal{A}$ as follows:
\begin{equation}
	\mathcal{A} = {\operatorname{argmax}}\ P(a|\mathcal{Q}, L_{src}\rightarrow L_{tgt}).
\end{equation}
Subsequently, the LLMs produce the result $\mathcal{R}$ by:
\begin{equation}
	\label{eq:res}
	\mathcal{R}_{tgt} = {\operatorname{argmax}}\ P(\mathcal{R}|\mathcal{A}, L_{tgt}).
\end{equation}
Finally, the LLMs determine the final results $\mathcal{Y}$ based on the reasoning path $\mathcal{R}_{tgt}$ in $L_{tgt}$ and generated alignment $\mathcal{A}$:
\begin{equation}
	\mathcal{Y} = {\operatorname{argmax}}\ P(y|\mathcal{A}, L_{tgt}, \mathcal{R}_{tgt}).
\end{equation}

\section{Automatic Cross-lingual Alignment Planning}
In this section, we introduce an Automatic Cross-lingual Alignment Planning (\textsc{AutoCAP}) framework to automatically select the most appropriate languages and weights for cross-lingual CoT, which consists of two main components: (1) \textit{Automatic Language Selection Prompting} and (2) \textit{Automatic Weight Allocation Prompting}.

\subsection{Automatic Language Selection Prompting}
To address the significant challenge of manual language selection, as shown in Figure~\ref{fig:framework}, we propose \textit{Automatic Language Selection Prompting} (ALSP) to autonomously and intelligently choose the most suitable languages and further utilize the cross-lingual capabilities of LLMs. Specifically, the prompt content is as follows:

\begin{mybox}
	\ \ 
	
	\textbf{Instruction:} ... your task is to select \Language{[Selected Number  $\mathcal{N}$]} languages optimal for cross-lingual reasoning on a given \SourceLanguage{[Source Language $\mathcal{L}_{src}$]} sample…\\
	\TargetLanguage{[Language Information $\mathcal{L}_{info}$]}\\
	\textbf{Input:}
	\InputName{[Given Query $\mathcal{Q}$]}\\
\end{mybox}

\noindent 
Specifically, ALSP directs LLMs to select the \Language{[Selected Number  $\mathcal{N}$]} of languages by analyzing the \InputName{[Given Query $\mathcal{Q}$]}, respective \SourceLanguage{[Source Language $\mathcal{L}_{src}$]}, and a comprehensive list of potential target language information \TargetLanguage{[Language Information $\mathcal{L}_{info}$]}. 
The language selection process can be formulated as:
\begin{equation}
	\mathcal{L}'_{tgt} \!= \!\underset{\mathcal{L}}{\operatorname{argmax}} \sum^\mathcal{N}_{i=1} P(\mathcal{L}^i_{tgt}|\mathcal{Q}, \mathcal{L}_{src}, \mathcal{L}^i_{info}),
\end{equation}
where $\mathcal{L}'_{tgt}= \{\mathcal{L}_{tgt}^{i'}\}^\mathcal{N}_{i=1}$ represents the final set of chosen target languages, and $\mathcal{L}^i_{info} \in \mathcal{L}_{info}$ encompasses respective language family, language branch, the proportion of available pre-training data to facilitate informed decision-making.

\subsection{Automatic Weight Allocation Prompting}

After selecting the target language, as shown in Figure~\ref{fig:framework}, we further introduce \textit{Automatic Weight Allocation Prompting} (AWAP). Specifically, our carefully designed prompts for guiding LLMs to automatically allocate weight to each language inference path are as follows:

\begin{mybox}
	\ \ 
	
	\textbf{Instruction:} After your language selection, please assign an alignment score for multilingual reasoning aggregation... \\
	\TaskName{[Weight Range $\mathcal{W}_{range}$]} \\
    \textbf{Input:} \InputName{[Given Query $\mathcal{Q}$]}\\

\end{mybox}

\begin{table*}
	\centering
	\begin{adjustbox}{width=\textwidth}
		\begin{tabular}{lccccccccccc}
			\toprule
			Model & bn & de & es & fr & ja & ru & sw & te & th & zh & AVG\\
			\midrule
			\rowcolor{gray!8}\multicolumn{12}{c}{\textit{GPT-3}$^\dagger$~\cite{brown2020language}} \\
			\midrule
			\texttt{Direct}~\cite{shi2022language} & 4.4 & 14.8 & 17.2 & 16.8 & 11.2 & 12.4 & 8.8 & 0.8 & 8.8 & 18.0 & 11.3 \\
			\texttt{Native-CoT}~\cite{shi2022language} & 6.4 & 36.0 & 40.4 & 37.6 & 26.0 & 28.4 & 11.2 & 0.4 & 10.8 & 40.0 & 23.7 \\
			\texttt{En-CoT}~\cite{shi2022language} & 9.6 & 44.0 & 44.8 & 46.0 & 32.4 & 28.4 & 20.8  & 5.6 & 19.6 & 40.8 & 29.2 \\
			\texttt{Translate-En}~\cite{shi2022language} & 41.2 & 46.4 & 51.6 & 46.4 & 44.8 & 48.8 & 37.6 & 42.8 & 41.2 & 47.2 & 44.8 
			\\
			\midrule
			\rowcolor{gray!8}\multicolumn{12}{c}{\textit{PaLM}$^\dagger$~\cite{chowdhery2023palm}} \\
			\midrule
			\texttt{Direct}~\cite{shi2022language} & 17.2 & 18.8 & 20.0 & 19.6 & 16.0 & 22.0 & 15.6 & 17.6 & 16.8 & 19.2 & 18.3\\
			\texttt{Native-CoT}~\cite{shi2022language} & 46.0 & 49.2 & 56.8 & 46.4 & 40.0 & 48.4 & 35.2 & 45.6 & 52.8 & 46.8 & 48.7 \\
			\texttt{En-CoT}~\cite{shi2022language} & 46.4 & 53.6 & 58.0 & 51.2 & 49.6 & 55.6 & 44.4 & 46.8 & 49.6 & 46.0 & 50.1 \\
			\texttt{Translate-En}~\cite{shi2022language} & 53.2 & 57.2 & 60.0 & 55.2 & 50.0 & 59.6 & 51.2 & 49.6 & 50.8 & 55.6 & 54.2 
			\\
			\midrule
			\rowcolor{gray!8}\multicolumn{12}{c}{\textit{GPT-3.5}~\cite{openai2022chatgpt}} \\
			\midrule
			\texttt{Direct$^\ddagger$}~\cite{qin2023cross} & 33.6 & 56.0 & 61.2 & 62.0 & 52.8 & 62.0 & 48.0 & 7.6 & 42.4 & 60.0 & 48.6 \\
			\texttt{Native-CoT$^\ddagger$}~\cite{qin2023cross} & 26.4 & 70.0 & 70.4 & 64.4 & 52.8 & 62.4 & 54.0 & 10.4 & 40.0 & 59.6 & 51.0 \\
			\texttt{En-CoT$^\ddagger$}~\cite{qin2023cross} & 50.0 & 73.6 & 69.6 & 70.0 & 60.4 & 65.6 & 55.2 & 22.0 & 48.0 & 63.2 & 57.8 \\
			\texttt{Translate-En$^\ddagger$}~\cite{qin2023cross} & 66.4 & 75.6 & 74.4 & 72.4 & 66.0 & 72.8 & 69.6 & \textbf{58.0} & 57.6 & 71.6 & 68.4 \\ 
			\texttt{CLP$^\ddagger$}~\cite{qin2023cross} & 64.8 & 80.0 & 82.4 & 79.2 & 69.2 & 81.6 & 74.8 & 38.8 & 62.0 & 73.6 & 70.6
			\\
			\texttt{CLSP$^\ddagger$}~\cite{qin2023cross} & 72.4 & 86.0 & 84.0 & 82.0 & 76.4 & 86.8 & 76.8 & 50.0 & 65.2 & 75.2 & 75.5 \\
			\texttt{Cross-ToT$^\ddagger$}~\cite{ranaldi2023empowering} & - & 87.6 & 86.2 & 84.3 & - & 86.5 & 75.4 & - & - & 83.5 & - \\  \hdashline
			\textsc{AutoCAP} & \textbf{76.0} & \textbf{88.0} & \textbf{86.8} & \textbf{84.4} & \textbf{79.6} & \textbf{88.0} & \textbf{78.4} & 52.0 & \textbf{69.2} & \textbf{84.0} & \textbf{78.6} \\

			\bottomrule
		\end{tabular}
	\end{adjustbox}
	\caption{
		Accuracy (\%) on MGSM. ``\texttt{Direct}'' prompt refers to directly asking and answering in the original language. ``\texttt{Native-CoT}'' prompt denotes answering with CoT in the native language. ``\texttt{En-CoT}'' prompt refers to mandating the use of CoT in English. ``\texttt{Translate-En}'' prompt signifies translating the query into English and then responding in English. The result with $^\dagger$ represents 6-shot sample prompt sourced from \citet{shi2022language}. The result with $^\ddagger$ indicates that it comes from \citet{qin2023cross} and \citet{ranaldi2023empowering}. For a fair comparison, for the integration method, we used 6 languages ​​for integration.
	}
	\label{main_results}
\end{table*}

\noindent
Specifically, this process dynamically allocates weights from \TaskName{[Weight Range $\mathcal{W}_{range}$]} to languages based on their relevance to the \InputName{[Given Query $\mathcal{Q}$]}, enhancing performance of LLMs by aligning the \SourceLanguage{[Source Language $\mathcal{L}_{src}$]} to target language $\mathcal{L}_{tgt}^{i'}$ generated from last turn more effectively.
Formally, the automatic weight for each language can be obtained by:
\begin{equation}
\mathcal{W}^{'}_i \!= \!\underset{w \in \mathcal{W}_{range}}{\operatorname{argmax}} p(w|\mathcal{Q}, \mathcal{L}_{src}\rightarrow \mathcal{L}_{tgt}^{i'}, \mathcal{L}^i_{info}),
\end{equation}
where $\mathcal{W}^{'}_i$ represents the cross-lingual alignment weight for the $i$-th target language.

\subsection{Automatic Cross-lingual Prompting Consistency}
By automatically determining the relevant language and its associated weight, our framework further adapts \textit{Automatic Cross-lingual Prompting Consistency} to more effectively merge multilingual alignments, leading to improved consistency across languages. Following Equation~\ref{eq:res}, we collect a set of generated results $\mathbf{R}$. The formulation of the final integrated result $\hat{\mathcal{R}}$ is presented as follows:
\begin{equation}
	\hat{\mathcal{R}} = \underset{\mathcal{R}\in \mathbf{R}}{\operatorname{argmax}} \sum^{\mathcal{N}}_{i=1}\sum_{r \in \mathbf{R}} \mathcal{W}^{'}_i \cdot \mathds{1} (\mathcal{R} = r),
\end{equation}
where $\mathcal{R}$ and $r$ both denote a reasoning outcome generated based on a specific formula in the target language $\mathcal{L}_{tgt}^{i'}$ from the generated result set $\mathbf{R}$, and $\mathcal{W}^{'}_i$ represents the weight assigned to that language. Additionally,  $\mathds{1}(X)$ is the indicator function, which returns 1 if $X$ is true and 0 if it is false.

\section{Experiments}
\subsection{Dataset and Baseline}

Following \citet{wei2022chain,qin2023cross}, we assess the performance of \textsc{AutoCAP} on the widely utilized multilingual mathematical reasoning dataset MGSM \cite{shi2022language} and select three representative LLMs as backbones, namely PaLM~\cite{chowdhery2023palm}, GPT3~\cite{brown2020language}, and GPT-3.5~\cite{openai2022chatgpt}. The top-p and temperature parameters in all processes are selected within the range of [0, 1]. 

\begin{table*}
	\centering
	\begin{adjustbox}{width=\textwidth}
		\begin{tabular}{lccccccccccl}
			\toprule
			Model & bn   & de   & es   & fr   & ja   & ru   & sw   & te   & th   & zh   & AVG  \\
			\midrule
			\textsc{AutoCAP} & 76.0 & 88.0 & 86.8 & 84.4 & 79.6 & 88.0 & 78.4 & 52.0 & 69.2 & 84.0 & 78.6 \\
			\midrule
			\ \ \textit{w/o AWAP} & 68.8 & 84.8 & 86.8 & 82.0 & 79.2 & 88.4 & 78.8 & 49.6 & 66.0 & 80.0 & 76.4 \textit{\textcolor{red}{(-2.2)}}\\
			\ \ \textit{w/o ALSP} & 60.0 & 80.4 & 84.8 & 81.2 & 73.6 & 85.6 & 64.8 & 45.2 & 66.8 & 78.0 & 72.0 \textit{\textcolor{red}{(-6.6)}} \\
			\ \ \textit{w/o AWAP \& ALSP} & 58.0 & 79.6 & 84.8 & 80.8 & 70.4 & 84.0 & 64.4 & 44.0 & 66.8 & 78.0 & 71.1 \textit{\textcolor{red}{(-7.5)}} \\ 
			\bottomrule
			\end{tabular}
	\end{adjustbox}
	\caption{Ablation experiment on GPT3.5. ``\textit{w/o ALWP}'' refers to removing \textit{Automatic Language Selection Prompting} (ALWP). ``\textit{w/o ALWP}'' refers to removing \textit{Automatic Weight Allocation Prompting} (ALSP). ``\textit{w/o ALWP \& ALSP}'' refers to removing both ALWP and ALSP.	}
	\label{ablation}
\end{table*}

\subsection{Main Results}

We follow previous work \citep{wei2022chain,qin2023cross} to adapt accuracy (Acc.) as the metric to evaluate the performance.
The main results are shown in Table~\ref{main_results}. 

From the results, we observe that \textsc{AutoCAP} attains superior performance compared to all baseline models by achieving state-of-the-art performance, even surpassing the ensemble methods \texttt{CLSP} and \texttt{Cross-ToT}, which manually select reasoning languages. Specifically, \textsc{AutoCAP} achieves an average accuracy improvement of over 3.1\%, outperforming \texttt{CLSP} across all tested languages. This demonstrates that \textsc{AutoCAP}, while implementing the automatic selection of reasoning languages, can better elicit the cross-lingual CoT reasoning capabilities of LLMs.

\subsection{Analysis}

To gain deeper insights into our approach, we explored the following research questions:
\vspace{0.2\baselineskip}

\noindent (1) \textit{Are all modules effective for \textsc{AutoCAP}?}

\noindent (2) \textit{Can the interactive capabilities of LLMs enhance the performance of \textsc{AutoCAP}?}

\noindent (3) \textit{Can \textsc{AutoCAP} work well on fewer  languages?}

\noindent (4) \textit{Why can \textsc{AutoCAP} works?}

\noindent (5) \textit{Can \textsc{AutoCAP} generalize to other open-source models?}

\noindent (6) \textit{Can \textsc{AutoCAP} generalize to other benchmarks?}

\noindent (7) \textit{What is the intuition behind \textsc{AutoCAP}?}

\subsubsection{Answer 1: All Modules in \textsc{AutoCAP} are Effective for \textsc{AutoCAP}}
In this section, we explore whether \textit{Automatic Language Selection Prompting} and \textit{Automatic Weight Allocation Prompting} are effective.
\paragraph{Automatic Language Selection Prompting is effective.}
To analyze the effectiveness for \textit{Automatic Language Selection Prompting} (ALSP), we removed the ALSP and randomly selected six languages for all query data instead.
As indicated in Table \ref{ablation} (\textit{w/o ALSP}), there is a significant decline in reasoning performance across all languages. In particular, there was an average accuracy reduction of 6.6\%.
It indicates that ALSP can effectively select the more optimal languages for alignment, which significantly improves the process of bridging linguistic gaps, directly contributing to a notable enhancement in the performance of multilingual CoT.

\paragraph{Automatic Weight Allocation Prompting is also crucial for performance enhancement.} 
To verify the impact of excluding the \textit{Automatic Weight Allocation Prompting} (AWAP) from our \textsc{AutoCAP}. By default, we adopt the setting of \citet{qin2023cross} and set the weight of all languages ​​to 1 by default.
As presented in Table \ref{ablation} (\textit{w/o AWAP}), it demonstrates a notable reduction in cross-lingual CoT performance, with an average decrease of 2.2\%.
The absence of specific weightings for each language resulted in lesser coherence when merging the outputs from multilingual reasoning, adversely affecting the reasoning process of overall effectiveness.
This performance decline highlights the critical role of the AWAP. Specifically, AWAP accurately allocates weights and facilitates detailed planning to enhance the degree of cross-lingual alignment, thereby improving the precision of multilingual reasoning.

\begin{figure}[t]
	\centering
	\includegraphics[width=0.49\textwidth]{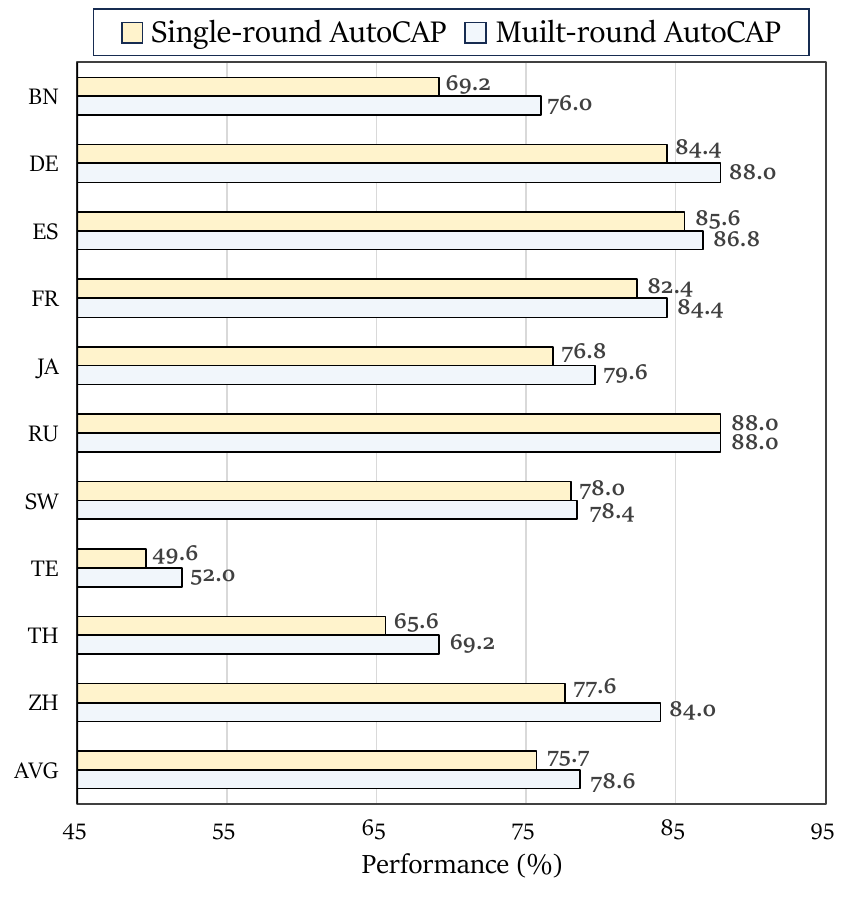}
	\caption{The accuracy of single-round \textsc{AutoCAP} compared to multi-round \textsc{AutoCAP}. AVG stands for average performance score.}
	\label{single}
\end{figure}

\paragraph{The combination of AWAP and ALSP brings a higher improvement.} 
To validate the functionality and effectiveness of the combination of the \textit{Automatic Weight Allocation Prompting} and \textit{Automatic Language Selection Prompting}, we conducted an experiment where both modules were simultaneously removed.
As shown in Table~\ref{ablation} (\textit{w/o AWAP \& ALSP}), we observe a significant decline in performance compared to the \textsc{AutoCAP}, with a decrease of 7.5\%. This decline was also evident when compared to the individual ablations of the ALSP and AWAP modules. The absence of both modules resulted in a substantial decrease in accuracy in multilingual reasoning, underscoring the importance of language selection and weighting for better cross-lingual reasoning.

\subsubsection{Answer 2: The interactive feature of LLMs boosts \textsc{AutoCAP} performance.}

To investigate the influence of the interactive capabilities of LLMs on \textsc{AutoCAP}, we differentiate between single-round prompting and multi-round prompting. Specifically, in the single-round prompting approach, we instruct LLMs to simultaneously select reasoning languages and allocate weights. Conversely, in the multi-round prompting approach (\textsc{AutoCAP}), in the first round, LLMs try to select the language, and then LLMs are required to allocate weight in the second round.

As illustrated in Figure~\ref{single}, the average performance of single-round interactions exhibited a decrease of 2.9\%, which indicates that leveraging the interactive capabilities of LLMs can significantly enhance the performance of cross-lingual CoT and the capability of language planning.

\subsubsection{Answer 3: \textsc{AutoCAP} also achieves positive results on fewer languages}
\begin{figure}[t]
	\centering
	\includegraphics[width=0.48\textwidth]{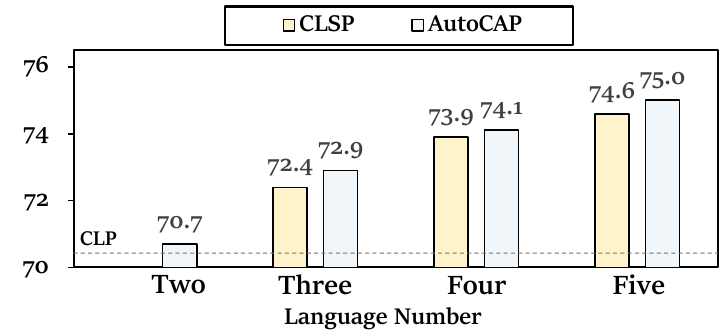}
	\caption{Performance comparison results of CLSP \cite{qin2023cross} and \textsc{AutoCAP} on different languages.
	}
	\label{number}
\end{figure}

To showcase the efficacy of our method using a limited number of languages, with a varied number of languages, specifically ranging from three to five. As shown in Figure~\ref{number}, a positive trend can be observed: as the number of languages incorporated increases, there is a corresponding enhancement in model performance. This pattern underscores the scalability and robustness of our approach in processing multilingual inputs.
Further, to explore the effectiveness of \textsc{AutoCAP}, we compared our results with the state-of-the-art method, CLSP \cite{qin2023cross}, which uses manually selected languages and a voting mechanism to merge answers from multiple languages. 
In detail, as shown in Figure~\ref{number}, when evaluated across linguistic settings encompassing three, four, and five languages, our model consistently outperformed the CLSP framework, registering an average performance uplift of 0.4\%. Such findings demonstrate the efficiency of our approach across varying language counts, which highlights the potential of our method in leveraging cross-lingual planning automatically to improve cross-lingual CoT capabilities.

\begin{figure}[t]
	\centering
	\includegraphics[width=0.48\textwidth]{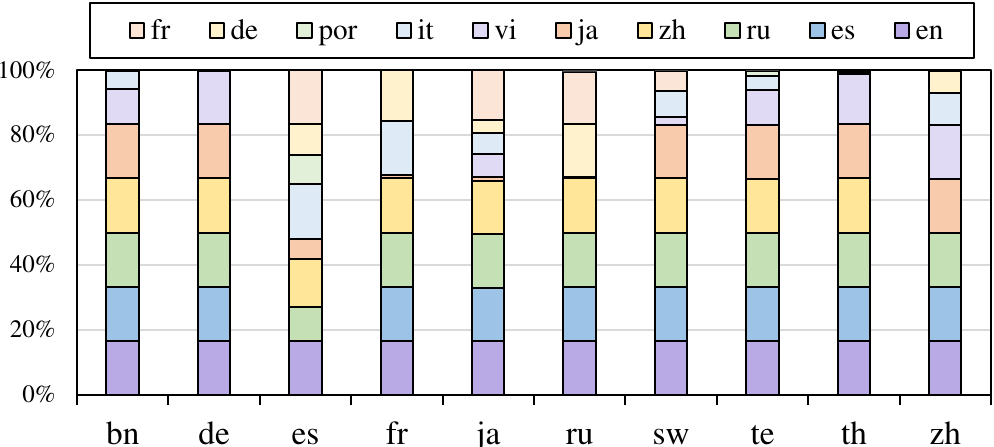}
	\caption{Languages for cross-lingual reasoning and their proportions.}
	\label{language}
\end{figure}

\begin{figure*}[t]
	\centering
	\includegraphics[width=160mm]{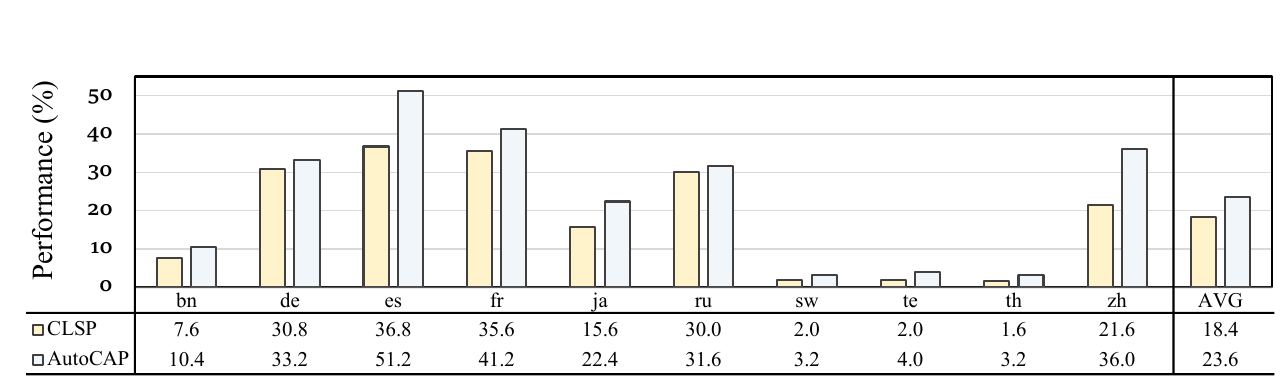}
	\caption{Comparison of the performance between \textsc{AutoCAP} and CLSP on the LLM Mistral \cite{jiang2023mistral}.}
	\label{mixtral}
\end{figure*}

\begin{figure}[t]
	\centering
	\includegraphics[width=0.49\textwidth]{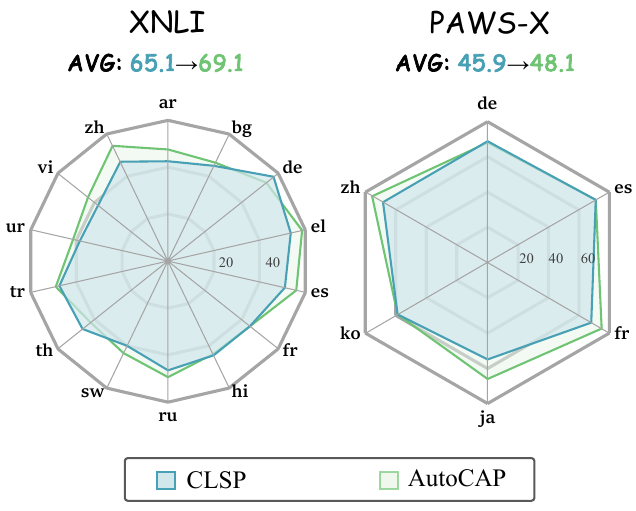}
	\caption{Accuracy on the XNLI \cite{conneau2018xnli} and PAWS-X \cite{yang2019paws} datasets.}
	\label{radar}
\end{figure}

\subsubsection{Answer 4: \textsc{AutoCAP} selects more diverse languages for better result}

In investigating the capacity of \textsc{AutoCAP} to encompass a broader linguistic spectrum, this section counts the variety and distribution of reasoning languages selected by \textsc{AutoCAP}. The statistical results are shown in the Figure~\ref{language}, in its reasoning processes, \textsc{AutoCAP} incorporated a minimum of seven and a maximum of ten distinct languages. This demonstrates a significant enhancement in linguistic diversity when contrasted with the conventional approach of static language selection, underscoring the superior adaptability and breadth of languages facilitated by the autonomous choices made by LLMs.

\subsubsection{Answer 5: \textsc{AutoCAP} also works on other LLM}
To further validate the scalability and universality of \textsc{AutoCAP}, we adapt \textsc{AutoCAP} on other open-source LLMs. The experimental outcomes on LLM Mistral \cite{jiang2023mistral} are depicted in Figure~\ref{mixtral}, demonstrating the optimal capability of \textsc{AutoCAP} on open-source LLMs. With an average performance improvement of at least 5.2\% over CLSP, these results further attest to the broad applicability of \textsc{AutoCAP}.

\vspace{0.7\baselineskip}

\subsubsection{Answer 6: \textsc{AutoCAP} exhibits strong generalization on other benchmarks.}

To further explore the effectiveness of \textsc{AutoCAP} on other tasks, following \citet{qin2023cross}, we conducted experiments on two multilingual datasets, XNLI \cite{conneau2018xnli} and PAWS-X \cite{yang2019paws}. The results, as shown in Figure~\ref{radar}, indicate that \textsc{AutoCAP} achieved better performance than CLSP on both datasets, with an average improvement of 4.0\% on XNLI and 2.2\% on PAWS-X. And it surpasses the performance of all languages compared with CLSP. These effectively illustrate the generalization of \textsc{AutoCAP} on different cross-lingual COT tasks.

\subsubsection{Answer 7: Qualitative analysis}

\begin{figure*}[t]
	\centering
	\includegraphics[width=160mm]{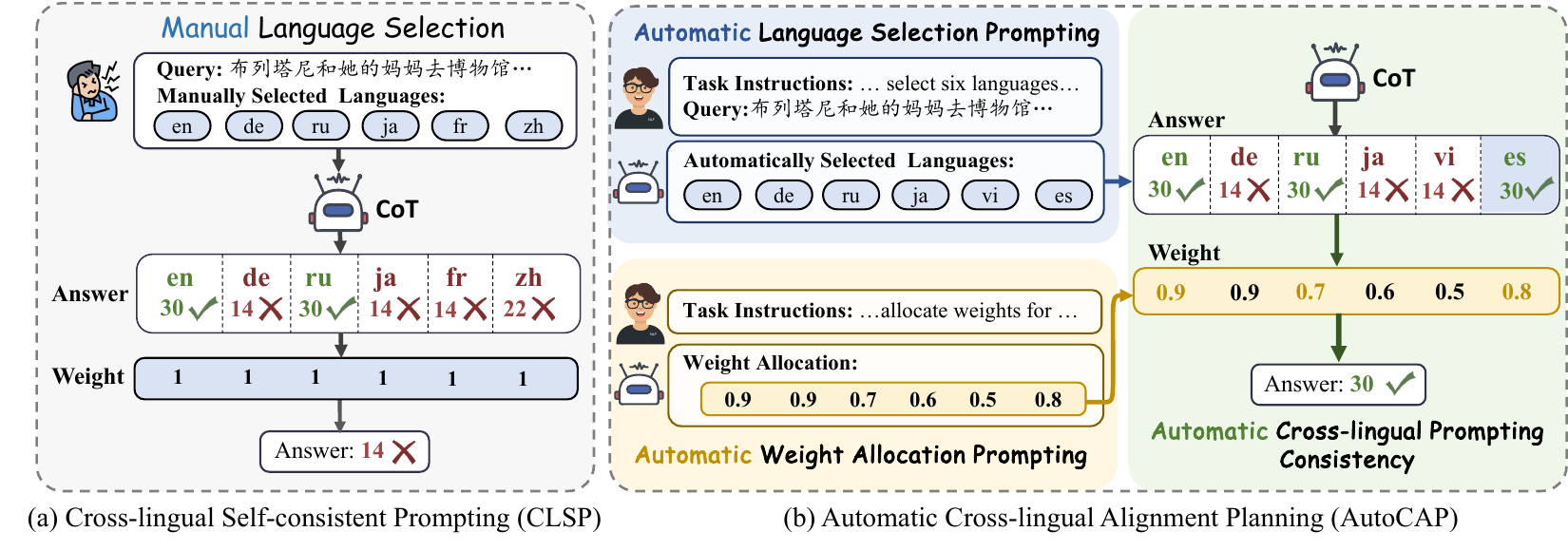}
	\caption{Case Study. Figure (a) illustrates the CLSP process, which integrates multiple reasoning answers through the manual selection of reasoning languages and a voting mechanism. Due to the predominance of the incorrect answer ``14'' in the results, this incorrect answer was selected. In contrast, Figure (b) demonstrates the ability of \textsc{AutoCAP} to automatically select aligned reasoning languages, with the correct inference answer ``30'' achieving a higher proportion than CLSP. By combining automatically allocated weights, the correct answer was selected.}
	\label{case}
\end{figure*}

To provide a more intuitive understanding of our method, we present a distinct case for qualitative analysis in this section. As shown in Figure~\ref{case}~(a), CLSP conducts reasoning in six manually selected languages. The reasoning results are correct in English (en) and Russian (ru), but incorrect in German (de), Japanese (ja), French (fr), and Chinese (zh). CLSP treats each reasoning path equally, integrating different paths solely through a voting mechanism, which unfortunately led to the incorrect answer ``14''.

\vspace{0.5\baselineskip}

Conversely, as depicted in Figure \ref{case}~(b), \textsc{AutoCAP} automatically selects six reasoning languages based on the query during the first interaction round. Although reasoning in German (de), Japanese (ja), and Vietnamese (vi) lead to incorrect answers, English (en), Russian (ru), and Spanish (es) produce correct ones.
In the second round, it assigns respective weight scores to each language. By aggregating these weighted scores, our method successfully circumvents the incorrect reasoning, ultimately selecting the correct answer ``30''.
These cases demonstrate the effectiveness and intuitiveness of our method. Specifically, our \textsc{AutoCAP} is capable of performing automatic cross-lingual planning on both languages and respective weights. Such advanced planning effectively decreases the cross-lingual alignment difficulties.

\section{Related Work}

The evolution of LLMs has profoundly propelled the progress in the field of artificial intelligence \cite{brown2020language,schaeffer2023emergent,yao2023tree,touvron2023llama,tang2023large}. In particular, CoT has provided a new perspective for solving complex problems and enhancing the reasoning capabilities of models by directing LLMs through a step-by-step problem-solving process.\cite{wei2022chain,kojima2022large,feng2023towards,zhang2023multimodal,zhang2022automatic}.

Current research primarily focuses on English, yet given the existence of over 7,000 languages worldwide, addressing critical challenges like reasoning and generation in minority languages has become an urgent necessity \cite{qin2023cross,shi2022language,lin2021few,huang2023not}.
Recognizing this gap, recent research has increasingly focused on transcending vanilla CoT to explore its cross-lingual dimensions.
In this vein, \citet{shi2022language} pioneer the introduction of the multilingual dataset specifically curated for mathematical reasoning. They advocate for a novel cross-lingual methodology that mandates LLMs to employ English in forecasting CoT sequences, translating the problem into English, and subsequently resolving it via English-based CoT paradigms. \citet{ranaldi2023empowering} unveil a self-consistent prompting mechanism within their cross-lingual, multi-step reasoning strategy, significantly augmenting reasoning capabilities across various languages. \citet{tanwar2023multilingual} suggest the integration of exemplars showcasing semantic congruence between source and target languages within the prompt context, aiming to facilitate seamless reasoning transitions across linguistic divides.
Moreover, \citet{qin2023cross} propose a self-consistent prompting, initially involving the manual selection of the reasoning language, followed by employing a voting mechanism to determine the final reasoning answer. This approach has yielded excellent results in the efficacy of cross-lingual prompting.

In comparison to the previous research, our work focuses on two key aspects. Firstly, we introduce an automatic language selection mechanism, enabling our system to choose the most accurately aligned reasoning languages for each query. Additionally, we develop an automatic weight allocation that effectively integrates the answers provided by various reasoning paths. To our knowledge, this is the first work to automatically select the reasoning languages and assign weights to reasoning paths.

\section{Conclusion}

In this paper, we present the Automatic Cross-lingual Alignment Planner (\textsc{AutoCAP}), a novel framework designed for enhancing zero-shot cross-lingual CoT reasoning. \textsc{AutoCAP} is comprised of two key components: \textit{Automatic Language Selection Prompting} and \textit{Automatic Weight Allocation Prompting}. 
\textsc{AutoCAP} achieves to automatically select suitable languages and allocate weights to various reasoning paths in cross-lingual CoT, respectively.
Extensive experiments demonstrate that \textsc{AutoCAP} achieves superior performance, outperforming existing methods in cross-lingual CoT.

\section{Limitations}

This work achieves automatic language selection and weight allocation for cross-lingual CoT, reducing manual workload significantly and marking a meaningful first step towards automatic resolution of multilingual tasks. In the future, we can apply our framework to multi-agent systems, enabling agents to incorporate automatic selection of reasoning languages and powerful tools to tackle challenging real-world problems.
In addition, we can consider the multi-lingual multi-modal CoT scenario by injecting the cross-modal CoT ability~\cite{chen2024m}.

\section{Acknowledgments}
This work was supported by the National Natural Science Foundation of China (NSFC) via grants 62306342, 62236004 and 62206078. This work was also sponsored by the Excellent Young Scientists Fund in Hunan Province (2024JJ4070). We are grateful for resources from the High Performance Computing Center of Central South University.

\bibliography{anthology,custom}
\bibliographystyle{acl_natbib}

\appendix

\end{CJK}
\end{document}